\documentclass[11pt,a4paper]{article}
\usepackage[hyperref]{acl2021}
\usepackage{times}
\usepackage{latexsym}
\usepackage{graphicx}
\usepackage{tikz}
\usepackage{float}
\usepackage{xspace}
\usepackage{booktabs}
\usepackage{threeparttable}
\usepackage{multirow}

\usepackage{microtype}

\aclfinalcopy 

\newcommand{\ours}{FeTaQA\xspace}

\def\checkmark{\tikz\fill[scale=0.4](0,.35) -- (.25,0) -- (1,.7) -- (.25,.15) -- cycle;}

\title{\textbf{\ours}: Free-form Table Question Answering}

\author{
        Linyong Nan$^{1}$
\quad   Chiachun Hsieh$^{3}$
\quad   Ziming Mao$^{1}$
\quad   Xi Victoria Lin$^{2}$\thanks{\ \ Now at Facebook AI.}
\quad   Neha Verma$^{1}$
\\{\bf
\quad   Rui Zhang$^{4}$
\quad   Wojciech Kry{\'s}ci{\'n}ski$^{2}$
\quad   Nick Schoelkopf$^{1}$
\quad   Riley Kong$^{5}$
\quad   Xiangru Tang$^{1}$}
\\{\bf
\quad   Murori Mutuma$^{1}$
\quad   Ben Rosand$^{1}$
\quad   Isabel Trindade$^{1}$
\quad   Renusree Bandaru$^{4}$}
\\{\bf
\quad   Jacob Cunningham$^{4}$
\quad   Caiming Xiong$^{2}$
\quad   Dragomir Radev$^{1,2}$}
\\
$^1$ Yale University 
\quad $^2$ Salesforce Research 
\quad $^3$ The University of Hong Kong
\\
\quad $^4$ Penn State University 
\quad $^5$ Archbishop Mitty High School
\\
\tt{\{linyong.nan, ziming.mao\}@yale.edu}, \tt{hsiehcc}@connect.hku.hk
}

\date{}

\begin{document}
\maketitle
\begin{abstract}
Existing table question answering datasets contain abundant factual questions that primarily evaluate the query and schema comprehension capability of a system, but they fail to include questions that require complex reasoning and integration of information due to the constraint of the associated short-form answers. To address these issues and to demonstrate the full challenge of table question answering, we introduce \ours, a new dataset with 10K Wikipedia-based $\textit{\{table}$, $\textit{question}$, $\textit{free-form answer}$, $\textit{supporting table cells\}}$ pairs.
\ours yields a more challenging table question answering setting because it requires generating free-form text answers after retrieval, inference, and integration of multiple discontinuous facts from a structured knowledge source. Unlike datasets of generative QA over text in which answers are prevalent with copies of short text spans from the source, answers in our dataset are human-generated explanations involving entities and their high-level relations.
We provide two benchmark methods for the proposed task: a pipeline method based on semantic parsing-based QA systems and an end-to-end method based on large pretrained text generation models, and show that \ours poses a challenge for both methods.
\end{abstract}

\section{Introduction}
Question Answering (QA) is the task of producing answers to natural language questions based on knowledge resources \cite{burke1997question,yao-van-durme-2014-information,chen-etal-2017-reading}.
One of the primary goals of QA is to allow users to directly and efficiently interact with large-scale and heterogeneous knowledge sources.
In the real world, knowledge sources take a variety of forms, including unstructured texts (documents, passages, or conversations), structured knowledge bases or databases, and semi-structured tables, each requiring dedicated modeling approaches.

For QA over text, a sequence modeling approach is usually adopted to encode the query and the context, and answers are either categorical \cite{lai-etal-2017-race}, extractive \cite{rajpurkar-etal-2016-squad, hotpotQA} or abstractive/generative \cite{kocisky-etal-2017-narrativeQA, nguyen-etal-2016-msmarco, fan-etal-2019-ELI5,natural2019kwiatkowski}.

For table-based QA, a common approach is to apply semantic parsing on the query and the table schema to generate a logical form (e.g. a SQL-like database query) that can be executed to retrieve the answer from the relevant portion of the table~\cite{pasupat-liang-2015-compositional, iyyer-etal-2017-search, zhong2017seq2sql, yu-etal-2018-spider}. The answers are extracted facts/entities in the table, therefore usually in short-form. 

\begin{figure*}[h!]
  \centering
  \includegraphics[width=1.0\textwidth]{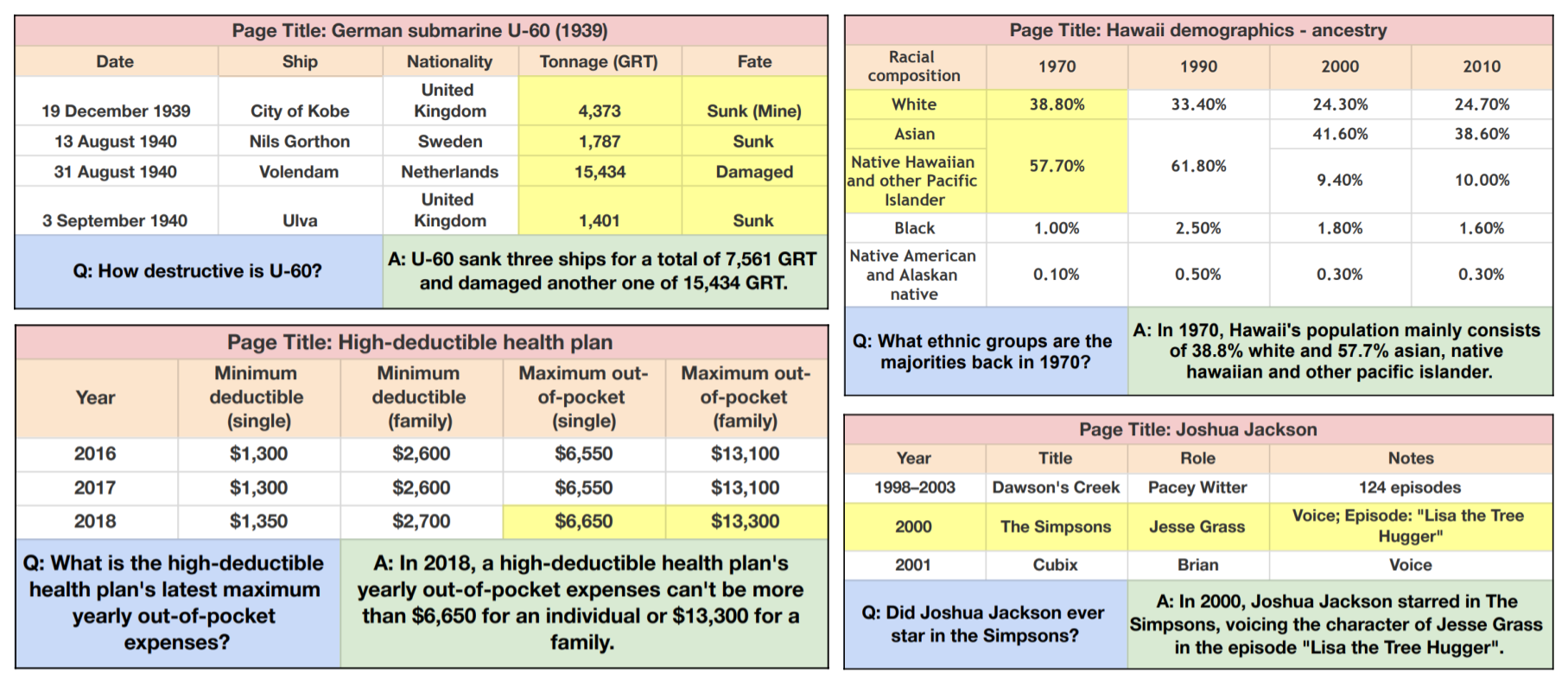}
  \caption{Examples of \textbf{\ours} instances. Only part of the original table is shown for better visualization. These examples are referred as (a), (b), (c), (d) from upper left to bottom right in the paper.}
  \label{fig:feta-instances}
\end{figure*}

\begin{table*}[]
\resizebox{\textwidth}{!}{%
\begin{tabular}{lcccccc}
\toprule
\multicolumn{1}{c}{\multirow{2}{*}{\textbf{Dataset}}} &  \multicolumn{4}{c}{\textbf{Knowledge Source}}                                                                             & \multicolumn{1}{c}{\multirow{2}{*}{\textbf{Answer Format}}} & \textbf{Avg \# Words} \\
\multicolumn{1}{c}{}    &  Wikipedia articles & Stories, books, movie scripts & Online forum texts & Wikipedia tables & \multicolumn{1}{c}{} &    \textbf{in Answer}       \\ \midrule
SQuAD \cite{rajpurkar-etal-2016-squad}   &     \checkmark         &                               &              &                  & Text-span       &        3.2                                    \\
HotpotQA \cite{hotpotQA}  &     \checkmark     &                               &              &                  & Short-form entity        &           2.2                         \\
NarrativeQA \cite{kovcisky2018narrativeqa}     &                    &        \checkmark                       &              &                  & Free-form text              &                4.7                \\
ELI5 \cite{fan-etal-2019-ELI5}   &                     &                               &       \checkmark       &                  & Free-form text               &           130.6                 \\
WikiTableQuestions \cite{pasupat-liang-2015-compositional}     &                    &               &              &         \checkmark         & Short-form entity           &              1.7                  \\
SequenceQA \cite{saha2018complex}   &                                  &                               &              &         \checkmark         & Short-form entity               &             1.2               \\
HybridQA \cite{chen2020hybridqa}   &  \checkmark                  &                               &              &         \checkmark         & Short-form entity                  &            2.1             \\ \midrule
\textbf{\ours}   &                    &                               &              &        \checkmark          & Free-form text                 &           18.9                  \\ \toprule
\end{tabular}
}
\caption{Comparison of \textbf{\ours} with other QA datasets.}
\label{tab:dataset-comparison}
\vspace{-3mm}
\end{table*}

Though existing datasets have enabled significant progress for table QA, their limitations prevent them from reflecting the full challenge of the task. Users of QA systems tend to ask complex questions which require elaborate answers, often containing explanations, while existing datasets are limited to simple short-form answers. 

To address these shortcomings we present \ours, a $\textbf{F}$r$\textbf{e}$e-form $\textbf{Ta}$ble Question Answering dataset which includes long, informative, and free-form answers. FeTaQA challenges QA systems with the following tasks: 1) retrieving multiple entities from tables based on the query; 2) reasoning over relations of these entities that are pertinent to the query and integrating these pieces of information into a coherent answer 3) aggregating  associated information into an explanation when the query is abstract or ambiguous; and 4) generating an informative, relevant, and faithful answer to the query. In addition, with tables sourced from Wikipedia, FetaQA covers a diverse set of topics and includes semi-structured tables containing un-normalized text, including numbers, dates, and phrases. \ours examples are presented in Figure \ref{fig:feta-instances} and differences between \ours and other QA datasets are described in Table~\ref{tab:dataset-comparison}.

We formulate 
generative 
table question answering as a Sequence-to-Sequence learning problem to evaluate the state-of-the-art models' performances on \ours. We propose two benchmark methods and provide experiment results for them. The first one is an end-to-end model that integrates query and table comprehension, logical reasoning, and language generation by adapting T5~\cite{raffel-2019-t5}. The other is a pipeline model that achieves content selection and surface realization in separate modules involving TAPAS~\cite{herzig-etal-2020-tapas}. 

Through human studies, we evaluate answers generated by our proposed models as well as the reference answer based on fluency, correctness, adequacy (informativeness), and faithfulness. 
The results indicate the challenging nature of \ours and that there is much room for improvement in QA systems.
We make the dataset available online.\footnote{\url{https://github.com/Yale-LILY/FeTaQA}}

\section{Dataset}
Here we introduce \ours
and describe the process and criteria for collecting the tables, questions and answers. Some statistics of \ours are shown in \S~\ref{sec:core_statistics}.

\subsection{Desiderata}
We frame 
generative
table question answering as the problem of 
generating an answer $a$ to a question $q$ based on a semi-structured table $T$ and its metadata $m$. 
Our goal was to construct a table QA dataset $\{(q_i, a_i, T_i, m_i)|i=1\ldots n\}$ that includes a large number of tables on diverse topics. The tables should be intelligible, well-formed, and moderately sized to make retrieval challenging yet plausible. Each table pairs a question with an answer sentence. The question should require retrieval and reasoning over multiple sources of information in the table, and the answer should integrate both facts and inferences into a coherent sentence that answers the question. Both questions and answers should be natural and fully grounded in the context of the entire table and its metadata such as the title.

\subsection{Data Collection Method}
We start building the dataset by collecting data instances from ToTTo \cite{parikh-etal-2020-totto}, a recent large-scale Table-to-Text dataset that contains tables and table-grounded sentences obtained from a diverse variety of Wikipedia pages. Additionally, ToTTo comes with annotations of table cells that support the sentence: a sentence is supported by the cell contents if it is directly stated or can be logically inferred by them. ToTTo applied several heuristics to sample the tables and the candidate sentences from Wikipedia pages, and their annotators are asked to revise sentences and highlight the corresponding table regions so that the sentences still have the varied language and structure found in natural sentences while being grounded to the table.

Sampling examples from the ToTTo dataset was conducted in multiple steps.
We first sample tables whose sizes are within 3 to 34 rows long and 3 to 7 columns wide (up to 75th percentile of all ToTTo table sizes)
to avoid truncation of sequence of linearized table for transformer-based models, whose default maximum input sequence length is 512. 
To ensure sentences contain several table entities, we further select tables whose annotation of highlighted regions covers multiple rows. 
We also collect a subset of single-row highlighted regions which span multiple rows or columns in content.
Following this sampling procedure, we were able to obtain 16,576 $\{\textit{table, metadata, highlighted region, sentence}\}$ instances with which we conduct the annotation procedure as described below. The flowchart of the sampling process is found in Figure \ref{fig:ToTTo_filtering_process} of the Appendix.

We adopted these table-grounded sentences as the answers in our new QA dataset since they are long, natural sentences containing rich information and inferences over the corresponding table. We also exploit ToTTo's annotations of table cells (the highlighted table region) as the 
weak supervision (denotations) 
for training models and labels for evaluating model retrieval competency. We parsed the tables (originally in HTML format) into a 2-dimensional array, where the first row corresponds to the table header. We also processed merged cells by copying the cell content and cell highlighted region to all the individual cells that compose the original merged cell.

\subsubsection{Question Annotation}
Question annotations were collected with the help of human judges in two phases: an internal phase conducted by on-site expert annotators, and an external phase conducted by crowd workers on Amazon Mechanical Turk. To streamline the process, we built a custom web interface to visualize table HTML and metadata, augmented with web widgets that allow table region highlighting and sentence editing. A screenshot of the annotation interface is shown in Figure \ref{fig:website-screenshot} of the Appendix.

Provided the necessary context, the annotators were asked to write a question 
whose answer is the provided ToTTo sentence.
The annotators were given the option to modify the sentence, the table cell content, and the highlighted region to better match the associated question. 
\paragraph{Internal Annotations}
In the first phase of annotation, we enrolled 15 internal annotators who were provided with preliminary guidelines. In addition to the annotation task, they were asked to provide feedback regarding the task instructions and the user experience of the website, based on which we iteratively modified the guideline and the website design.

\paragraph{External Annotations}
For external annotations, we hired MTurk workers who have completed at least 500 HITs, have 97\% approval rate, and are from English-speaking regions. To ensure that the MTurk annotators understand our task, we provided an instruction video for the interactive annotation tool usage, FAQs that clarify the annotations we desire, along with good vs. bad annotation examples. 
We also created a Slack channel for crowdsourced workers to ask questions and clarify doubts. 

\paragraph{Annotation Evaluation}
To ensure \ours is of high quality, we evaluate crowdsourced annotations as follows. 
First we auto-rejected questions that fall outside the length range (4 to 25) or convoluted questions that contain more than two interrogatives (259 examples in total).
For the remaining annotations, we built another web interface for evaluation and asked internal evaluators to label an annotation as ``approve'', ``reject'' or ``questionable'' and score the annotation based on its fluency, faithfulness, and the extent to which the question needs the full sentence as the answer. 
Internal evaluators were also asked to modify the question annotations that were not approved.
Our final dataset annotators contribution is distributed as follows: we have 3,039 (30\%) instances from internal annotators, 7,291 (70\%) from MTurk workers. In total, our dataset contains 10,330 instances.

\subsection{Dataset Split}
Randomly splitting the dataset may make train, development, and test splits contain tables with similar contents~\cite{dollak2018improving,lewis2020question}.
Therefore, to increase the generalization challenge, we calculated the Jaccard similarity of two instances based on the set of tokens shown in table headers and questions, and split the dataset in such a way that models are evaluated on test split instances that are least similar to those used for training. We first sampled 800 instances randomly as a seed split. Then we add those that have Jaccard similarities greater than 0.465 to the seed split. This process generates two splits of 70\% and 30\% of all instances, the former becomes the train split and the latter is randomly divided with a ratio of 1:2 to form the development and test splits. This results in 7,326/1,001/2,003 instances in the train/dev/test splits, respectively.

\subsection{Data Analysis and Statistics}
\label{sec:core_statistics}
Basic statistics of \ours are shown in Table \ref{tab:basic-stats}, and human evaluation scores and inter-evaluator agreements are reported in Table \ref{tab:agreement}. A quantitative and qualitative analysis of \ours shows it contains abundant complex questions that require retrieval of multiple entities in the context, as shown by the human evaluation score for question complexity, and that the median number of highlighted cells (denotations) is 6, which is twice as much as the corresponding number for ToTTo. These denotations are correct and adequate as indicated by the corresponding high evaluation scores. The free-form answers have a median of 18 tokens in length, and are grounded to the table and the denotations, also suggested by the high evaluation scores.

\begin{table}[ht!]
\centering 
\resizebox{.45\textwidth}{!}{%
\begin{tabular}{lr}
\toprule
\textbf{Property}                            & \textbf{Value}  \\
\midrule
Unique Tables                                & 10,330          \\
Question Length (Median/Avg)                 & 12 / 13.2       \\
Answer Length (Median/Avg)                   & 18 / 18.9       \\
Rows per Table (Median/Avg)                  & 12 / 13.8       \\
Columns per Table (Median/Avg)               & 5 / 5.9         \\
No. of Highlighted Cell (Median/Avg)         & 6 / 8.0         \\
Percentage of Cells Highlighted (Median/Avg) & 10.7\% / 16.2\% \\
Page Title Length (Median/Avg)               & 2 / 3.3         \\
Section Title Length (Median/Avg)            & 2 / 1.9         \\
\midrule 
Training Set Size                            & 7,326           \\
Development Set Size                         & 1,001           \\
Test Set Size                                & 2,003           \\
\toprule
\end{tabular}
}
\caption{\ours Core Statistics}
\label{tab:basic-stats}
\end{table}

\begin{table}[ht!]
\centering
\resizebox{\columnwidth}{!}{%
    \begin{tabular}{lccc}
    \toprule
    \textbf{Annotation Quality} & \textbf{\begin{tabular}[c]{@{}c@{}}Score \textgreater{}= 4 \\ (\%)\end{tabular}} & \textbf{\% Agreement} & \textbf{\begin{tabular}[c]{@{}c@{}}Randolph's Kappa \\ / 95\% CI\end{tabular}} \\ \midrule
    Question Complexity         & 52.6                                  & 0.65                  & 0.48 / {[}0.41, 0.55{]}             \\
    Denotation Correctness      & 89.0                                  & 0.88                  & 0.82 / {[}0.76, 0.88{]}             \\
    Denotation Adequacy         & 91.6                                  & 0.89                  & 0.83 / {[}0.77, 0.89{]}             \\
    Answer Fluency              & 95.0                                  & 0.92                  & 0.89 / {[}0.84, 0.94{]}             \\
    Answer Correctness          & 92.4                                  & 0.91                  & 0.86 / {[}0.80, 0.92{]}             \\
    Answer Adequacy             & 90.6                                  & 0.88                  & 0.82 / {[}0.76, 0.88{]}             \\
    Answer Faithfulness         & 95.6                                  & 0.93                  & 0.89 / {[}0.84, 0.94{]}             \\ \toprule
    \end{tabular}
}
\caption{Human evaluation over 100 samples of \ours. 5 internal evaluators are asked to rate the samples on a scale of 1 to 5. We report $\%$ of samples that have score $\geq 4$ to show high quality of \ours, and report percent agreement and Randolph's Kappa \cite{Randolph2005FreeMarginalMK} (with 95$\%$ CI) to show that our human evaluation has high inter-annotator agreement.
}
\label{tab:agreement}
\end{table}

\begin{figure}[ht!]
  \centering
  \includegraphics[width=0.45\textwidth]{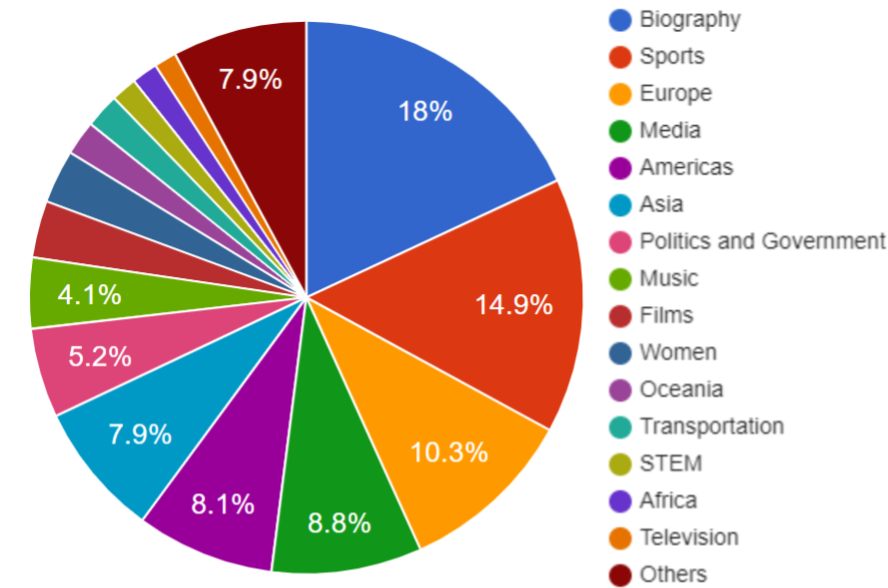}
  \caption{\ours Topics Distribution.}
  \label{fig:topic-distribution}
\end{figure}

\paragraph{Topics} 
Similar to ToTTo, we use Wikimedia Foundation's topic categorization model 
\cite{wikimedia-topics} to investigate the topics distribution of \ours. Although our dataset is limited to topics presented in ToTTo, we are able to sample instances that have evenly distributed topics, as shown in Figure \ref{fig:topic-distribution}. We found that most of the instances are related to biography, sports and geographical regions. There are also abundant instances related to media, politics and government. 

\begin{figure*}[ht!]
  \centering
  \includegraphics[width=\textwidth]{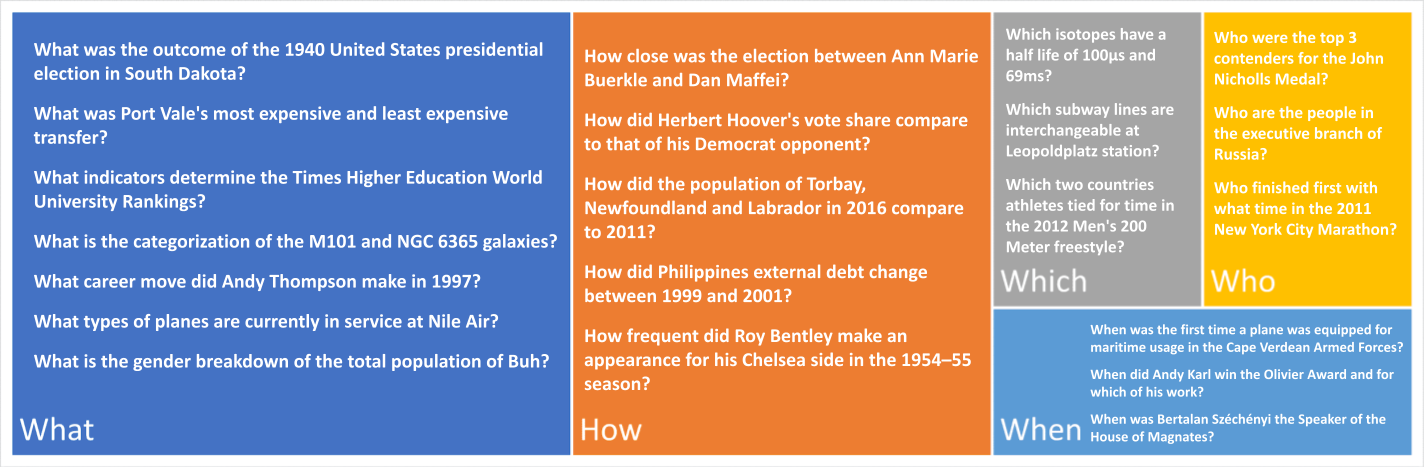}
  \caption{\ours questions by top 5 most frequent starting words, where box size represents frequency.}
  \label{fig:fetaqa-question-type-treemap}
\end{figure*}
\paragraph{Question Types}
\ours has diverse and complex questions, as illustrated in Figure \ref{fig:fetaqa-question-type-treemap}. Comparison of question type distributions with other table QA datasets is shown in Figure \ref{fig:all-question-types} of the Appendix.
We found that in \ours, a large percentage of $\textit{what}$ questions are asking entities in plural, or abstract entity such as $\textit{outcome, result, margin, percentage}$. In addition, there is a higher percentage of $\textit{how}$ questions that are not $\textit{how many/much}$, compared to existing table QA datasets.

\section{Models}
\begin{figure*}[h!]
  \centering
  \includegraphics[width=0.9\textwidth]{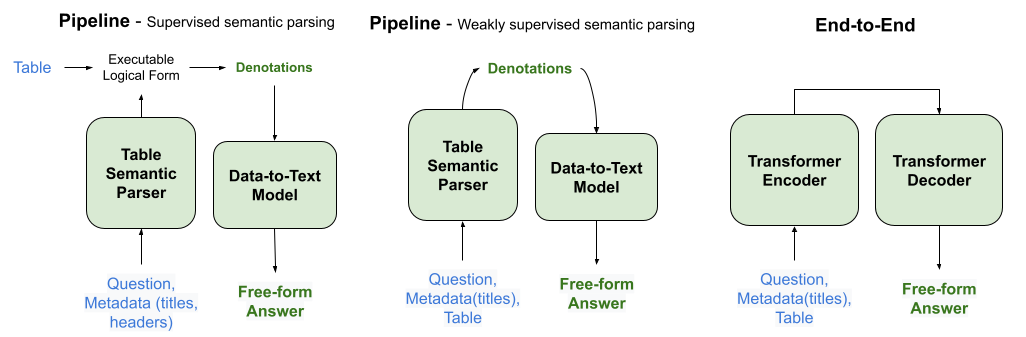}
  \caption{Pipeline model and End-to-End model diagrams.}
  \label{fig:model-diagrams}
\end{figure*}

To quantify the 
challenge posed by \ours for state-of-the-art models, we used two modeling approaches that 
have been shown to be effective for the existing table question answering datasets, with some modifications made to adjust to our task. Model configurations are shown in Figure \ref{fig:model-diagrams}.

\subsection{Pipeline Model}
Question answering over tables is usually seen as a semantic parsing task. Based on the table schema, a semantic parser maps the question to a logical form that can be used to retrieve the result from the table. The answer is usually a single entity that is either a table cell value or the aggregation result of multiple cell values (aggregation operators include COUNT, MAX, SUM, etc.). A table semantic parser is trained using logical forms as supervised examples, but due to its high annotation cost, an alternative approach is to use denotations for weak supervision. The denotations are usually the targets of the existing table QA tasks. With this approach, the parser is able to retrieve denotations directly.

However, in our task, targets are generated texts instead of retrieved denotations, suggesting that we also need a generator to integrate the retrieved information into a cogent sentence.
Therefore, we propose a pipeline model with two separately trained modules, described below. 

\paragraph{Weakly Supervised Table Semantic Parsing}
The first module adopts a table semantic parser that is pre-trained with weak supervision. We use TAPAS \cite{herzig-etal-2020-tapas}, a state-of-the-art model for table QA, to start with. We fine-tune it on \ours with our annotated denotations (highlighted table regions). We believe fine-tuning is crucial for our task because TAPAS is pre-trained on questions that require retrieval of limited denotations (single entity or homogeneous entities that can be aggregated with $\texttt{COUNT}$, $\texttt{SUM}$, or $\texttt{AVG}$ operation), while \ours questions require retrieval of multiple entities and complex aggregation operations. Details of experiment results are provided in Section \ref{sec:results}.
Note that besides denotations, TAPAS also predicts an aggregation operation (choose from \texttt{COUNT}, \texttt{SUM}, \texttt{AVG}, \texttt{NONE}) applied to the predicted denotations to obtain the final answer. However, we use $\texttt{NONE}$ as the aggregation operation label for fine-tuning due to the lack of annotations, therefore leaving the inference of aggregation operation to the second module.  

\begin{figure}[h]
  \centering
  \includegraphics[width=0.45\textwidth]{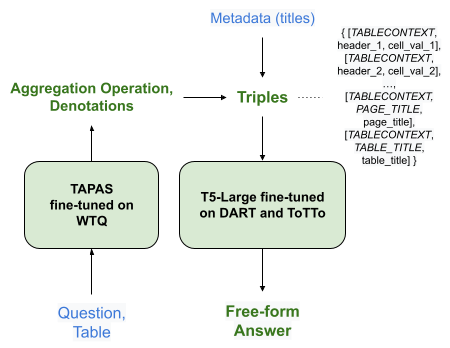}
  \caption{Weakly supervised fine-tuning of table semantic parser on \ours. We choose a checkpoint of TAPAS-base fine-tuned on WikiTableQuestions to start with. After fine-tuning, the table semantic parser predicts denotations, which are then converted to triples and sent to the Data-to-Text module.}
  \label{fig:pipeline-finetuning-detail}
\end{figure}

\paragraph{Data-to-Text}
\label{sec:data-to-text}
As shown in Figure \ref{fig:pipeline-finetuning-detail}, we fine-tune T5 \cite{raffel-2019-t5} on DART \cite{dart} to obtain a Data-to-Text model as the second module of the pipeline to perform surface realization of table cells (denotations in our case). We first convert the denotation prediction into the triple-set format with the following scheme: for each table cell in the highlighted region, we generate the following triple: $\big[\texttt{[TABLECONTEXT]}$, $\texttt{column\_header}$, $\texttt{cell\_value} \big]$, where $\texttt{column\_header}$ is the cell's corresponding column name. Similar to DART, we use $\texttt{[TABLECONTEXT]}$ as a special token for converting a table cell into a triple. We then incorporate the metadata into triples by replacing $\texttt{column\_header}$ with the field name ($\texttt{TABLE\_TITLE}$, $\texttt{PAGE\_TITLE}$) and $\texttt{cell\_value}$ with the metadata content (table title text, page title text).
We end up with a triple-set containing all highlighted table cells and the metadata (table title and title of the Wikipedia page that includes the table). We further fine-tune the Data-to-Text model on ToTTo 
instances so that it adapts to our formation of triple-set inputs. 
To avoid exposure to \ours test instances, we fine-tune with a sample of 8K ToTTo instances that are not used for creating \ours.  
 
\subsection{End-to-End Model}
\begin{figure}[h]
  \centering
  \includegraphics[width=0.45\textwidth]{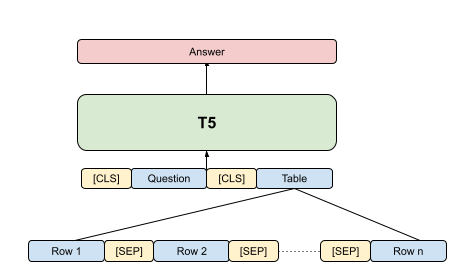}
  \caption{Table linearization in end-to-end model.}
  \label{fig:end2end-finetune-detail}
\end{figure}

In this approach, we model the task as a sequence-to-sequence learning problem by linearizing table $T$ appended to question $q$ as the source sequence, and treating the free-form answer $a$ as the target sequence. We propose a simple linearization scheme as a baseline: table rows are concatenated with $\texttt{[SEP]}$ tokens in between, and cells in each row are separated by spaces. Since the input sequence length may exceed the model limit, we prepend $q$ to table linearization $\widetilde{T}$, using $\texttt{[CLS]}$ tokens as prefixes for separation. We fine-tune models from the T5-family on the \ours train set. The linearization scheme is visualized in Figure \ref{fig:end2end-finetune-detail}.

\section{Experiments}
In this section, we explain the experiment settings and report the automatic and human evaluations on model outputs.

\subsection{Experiment Setup}
We first experiment with the pipeline model in a zero-shot setting, that is, without any fine-tuning on \ours. We use a checkpoint of TAPAS-base that is fine-tuned on WikiTableQuestions \cite{pasupat-liang-2015-compositional} to perform table semantic parsing implicitly in order to produce a set of denotations, which is then converted to a triple-set as described in \ref{sec:data-to-text}. We then employ a T5-large model \cite{raffel-2019-t5} that goes through two fine-tuning stages: in the first stage it is fine-tuned on the downstream Data-to-Text task with DART \cite{dart};
in the second stage it is further fine-tuned on ToTTo instances to adapt to the triple-set formulation we proposed. 
We denote this setting as $\texttt{Pipeline - zeroshot}$ in Table \ref{tab:experiment-results}. 
Next we experiment with the pipeline model by fine-tuning the table semantic parser on \ours. We further fine-tune the TAPAS-base checkpoint (WTQ fine-tuned) on \ours train set and select models based on their performance on the development set. We use the same Data-to-Text model as described in the zero-shot setting.

For the End-to-End model, we adapt
Hugging Face's implementation \cite{wolf-etal-2020-transformers} of T5 \cite{raffel-2019-t5} for our task. We use a standard T5-tokenizer with additional $\texttt{[CLS]}$ and $\texttt{[SEP]}$ tokens and the model vocabulary is resized accordingly. Since we expect the input sequence to be significantly longer than the target, we fine-tuned the models using T5's ``summarize: " prefix. The motivation behind this is to avoid simple extraction from the table since abstractive summarization is supposed to rephrase important details in the source.
T5-small is trained on 4 Tesla K80 GPUs with per-device batch size of 16 for 30 epochs (about 6,900 steps). T5-base is trained on 4 Tesla K80 with per-device batch size of 4 (due to GPU memory constraints) for 80 epochs (about 36,640 steps). As for T5-large, we distributed the layers across 8 Tesla K80 to train with a batch size of 4 for 80 epochs (about 80k steps).

\subsection{Evaluation Metrics}
We use a variety of automatic metrics and human evaluation (Section \ref{sec:human-eval}) to evaluate the quality of the generated answers. We report sacreBLEU \cite{post-2018-call}, ROUGE-$\{$1, 2, L$\}$ \cite{lin-2004-rouge}, and METEOR \cite{banerjee-lavie-2005-meteor} that evaluate the n-gram match between generated and reference answers. Considering the limitations of these measures in evaluating the semantic meanings of sentences, we also report BERTScore \cite{zhang2020bertscore} and BLEURT \cite{sellam2020bleurt} that incorporate semantics using contextual embeddings.
To evaluate the retrieval competency of table semantic parsers, we applied various set similarity metrics to the predicted and reference denotation lists. Specifically, we report Jaccard similarity, Overlap, Cosine similarity, and Dice similarity.

\subsection{Results and Discussions}
\label{sec:results}

\begin{table*}[]
\centering
\resizebox{\textwidth}{!}{%
    \begin{tabular}{lccccccc}
        \toprule
                                               & \textbf{sacreBLEU}\footnotemark[1]      & \textbf{ROUGE-1}       & \textbf{ROUGE-2}       & \textbf{ROUGE-L}       & \textbf{METEOR}        & \textbf{BERTScore}     & \textbf{BLEURT}        \\ 
        \midrule
        Pipeline - zeroshot                     & 9.16           & 0.38          & 0.20          & 0.33          & 0.22          & 0.88          & -0.79         \\
        Pipeline - fine-tuned & 11.00          & 0.40          & 0.22          & 0.35          & 0.24          & 0.91          & -0.35         \\
        \midrule
        End-to-End - T5-small                   & 21.60          & 0.55          & 0.33          & 0.47          & 0.40          & 0.94 & 0.08 \\
        End-to-End - T5-base                    & 28.14          & 0.61          & 0.39          & 0.51          & 0.47          & 0.96          & 0.31         \\
        End-to-End - T5-large                   & \textbf{30.54} & \textbf{0.63} & \textbf{0.41} & \textbf{0.53} & \textbf{0.49} & \textbf{0.96} & \textbf{0.57}          \\
        \toprule
    \end{tabular}
}
\caption{Experiment results on the test split of \ours.}
\label{tab:experiment-results}
\end{table*}

Our experimental results on the \ours test set are summarized in Table \ref{tab:experiment-results}. The T5-large model using an End-to-End modeling approach achieves the highest performance scores in almost all evaluation metrics. Also, we observe a large performance gap between pipeline models and End-to-End models, even though the latter only adopt a simple linearization strategy for encoding tables.

We also see that after fine-tuning on \ours with denotations as weak supervisions, the pipeline model improves by almost 2 BLEU points. To further examine the source of this improvement, we report the evaluation of table semantic parser performance in Table \ref{tab:denotation-results}, from which we also observe an improvement in retrieval capability. However, we note that compared with the reference denotations that have a median of six table cells being highlighted (shown in \ref{tab:basic-stats}), our table semantic parser is only able to predict two table cells on average before fine-tuning on \ours, and three table cells on average after. This indicates a large space for improvement. We suspect that the low performance of denotation predictions and the loss of relational information between denotations lead to the inadequate performance of pipeline models.

\begin{table}[]
\centering
\resizebox{.45\textwidth}{!}{%
\begin{tabular}{ccccc}
\toprule
           & \textbf{Jaccard} & \textbf{\begin{tabular}[c]{@{}c@{}}Overlap\\ Coff.\end{tabular}} & \textbf{Cosine} & \textbf{Dice} \\ \midrule
Zeroshot   & 0.065            & 0.300                                                            & 0.140           & 0.109         \\
Fine-tuned & 0.101            & 0.311                                                            & 0.184           & 0.161         \\ \toprule
\end{tabular}
}
\caption{Evaluation of denotation prediction on the test split of \ours. We report performance of TAPAS in zero-shot and fine-tuned with weak supervision.}
\label{tab:denotation-results}
\end{table}

\subsection{Human Evaluation}
\label{sec:human-eval}
To further evaluate the quality of the answers generated by different models comparing to the references, we conduct our human evaluation based on four criteria: (1) $\textit{fluency}$ if an answer is natural and grammatical; (2) $\textit{correctness}$ if an answer is correct; (3) $\textit{adequacy}$ if an answer contains all the information that is asked; (4) $\textit{faithfulness}$ if an answer is faithful and grounded to the contents of the table and the highlighted region. Each evaluator is asked to examine an answer given the question and the full context (table, highlighted region, and metadata) and give a score on a scale of 1 to 5 for each of the criteria. We ask five internal annotators to evaluate 100 samples of \ours instances. Each sample is paired with 3 answers: the reference, the pipeline model result, and the End-to-End model result.

Table \ref{tab:human-eval} attests to the high quality of our annotations and the challenging nature of \ours. Similar to the evaluation result of the automatic metrics, we observe a large gap between the pipeline model and the End-to-End model, with the latter one significantly outperforming its counterpart in terms of answer correctness, adequacy, and faithfulness. Comparing the best performing End-to-End model outputs to human references, we see that there is room for improvement in the future. 

\begin{table}[]
\centering
\resizebox{0.5\textwidth}{!}{%
\begin{tabular}{ccccc}
\toprule
\textbf{Source} & \textbf{\begin{tabular}[c]{@{}c@{}}Fluent\\ (\%)\end{tabular}} & \textbf{\begin{tabular}[c]{@{}c@{}}Correct\\ (\%)\end{tabular}} & \textbf{\begin{tabular}[c]{@{}c@{}}Adequate\\ (\%)\end{tabular}} & \textbf{\begin{tabular}[c]{@{}c@{}}Faithful\\ (\%)\end{tabular}} \\ \midrule
Pipeline        & 85.2                                                             & 25.4                                                              & 8.4                                                               & 23.6                                                               \\
End-to-End      & 94.6                                                             & 54.8                                                              & 48.4                                                               & 50.4                                                               \\
Reference       & 95.0                                                            & 92.4                                                              & 90.6                                                               & 95.6                                                               \\ \toprule
\end{tabular}
}
\caption{Human evaluation over 100 samples of model outputs and references. We report the percentage of outputs that have scores of 4 or 5.}
\label{tab:human-eval}
\end{table}

\section{Related Work}
\paragraph{Generative QA} Generative question answering datasets such as NarrativeQA~\cite{kovcisky2018narrativeqa},
CoQA~\cite{reddy2019coqa}, 
TriviaQA~\cite{joshi2017triviaqa}, and MS MARCO~\cite{nguyen-etal-2016-msmarco} all have free-form answers that are generated based on the contexts of Wikipedia articles, books, movie scripts, dialogues or web documents. These responses are mostly crowd-sourced and are reported to mostly contain copies of short text spans from the source.
By contrast, ELI5~\cite{fan-etal-2019-ELI5} is a long form question answering dataset containing a diverse set of complex questions, each paired with a paragraph-long answer and 100 relevant $\textit{web source}$ documents~\cite{petroni2020kilt,krishna2021lfqa}. 
\ours is the first dataset for generative question answering over tables. Unlike the existing generative QA datasets that assess multi-documents retrieval and abstraction capability, \ours poses new challenges in the reasoning and integration capability of a system given a structured knowledge source. 

\footnotetext[1]{SacreBLEU signature:\\ BLEU+case.lc+numrefs.1+smooth.exp+tok.13a+version.1.3.7}

\paragraph{QA over Tables and Semantic Parsing} 
Several datasets have been proposed to apply semantic parsing on tables, including WikiTableQuestions~\cite{pasupat-liang-2015-compositional}, SequentialQA~\cite{iyyer-etal-2017-search}, WikiSQL~\cite{zhong2017seq2sql}, Spider~\cite{yu-etal-2018-spider}. With the development of pre-trained language models, recent work~\cite{yin-etal-2020-tabert,herzig-etal-2020-tapas,eisenschlos-etal-2020-understanding,iida2021tabbie} jointly learns representations for natural language sentences and structured tables, and \citet{yu2020grappa,yu2021score} use pre-training approach for table semantic parsing. HybridQA~\cite{chen2020hybridqa} and OTT-QA~\cite{Chen2020OpenQA} have contexts of both structured tables and unstructured text. MultiModalQA~\cite{talmor2021multimodalqa} contains complex questions over text, tables and images. These datasets define a table QA task that is extractive in nature by restricting their answers to be short-form, while \ours frames table QA as a generation task.

\paragraph{Data-to-text generation}
Recent neural end-to-end models tested on the WebNLG 2017 dataset \cite{gardent-etal-2017-webnlg} have focused on incorporating pre-training and fine-tuning for specific generation tasks~\cite{Chen2020KGPTKP,kale2020text} to improve performance and strengthen generalization ability. 
However, recent models featuring separate content-planning and surface realization stages have exhibited improvements ~\cite{moryossef2019stepbystep,iso2020learning} over comparable baselines. 
TabFact~\cite{Chen2020TabFactAL} is composed of Wikipedia tables coupled with statements labeled as either ``ENTAILED'' or ``REFUTED'' by the table. LogicNLG~\cite{Chen2020LogicalNL} features statements logically entailed from tables. 
ToTTo~\cite{parikh-etal-2020-totto} is a large-scale open-domain dataset consisting of Wikipedia tables with a set of highlighted table cells and a sentence description of those highlighted cells. DART~\cite{dart} is an open-domain Data-to-Text dataset that contains table-ontology-preserving data samples with diverse predicate set occurred in Wikipedia tables.

\section{Conclusion}
In this paper, we introduced the task of 
generative table question answering with \ours, a table QA dataset consisting of complex questions that require free-form, elaborate answers. We also proposed two modeling approaches: (1) a pipeline model that incorporates a table semantic parser and (2) a Data-to-Text generator, and an End-to-End model that includes query comprehension, reasoning and text generation. Our experimental results indicate that the End-to-End model with a simple table encoding strategy achieves much higher scores than the pipeline model that requires table semantic parsing. Furthermore, we show that \ours introduces new challenges for table question answering that call for innovative model designs in the future.

\bibliographystyle{acl_natbib}
\bibliography{anthology,acl2021}

\clearpage
\newpage
\appendix
\section{Appendix}
\label{sec:appendix}

\begin{figure*}[ht]
  \centering
  \includegraphics[width=0.9\textwidth]{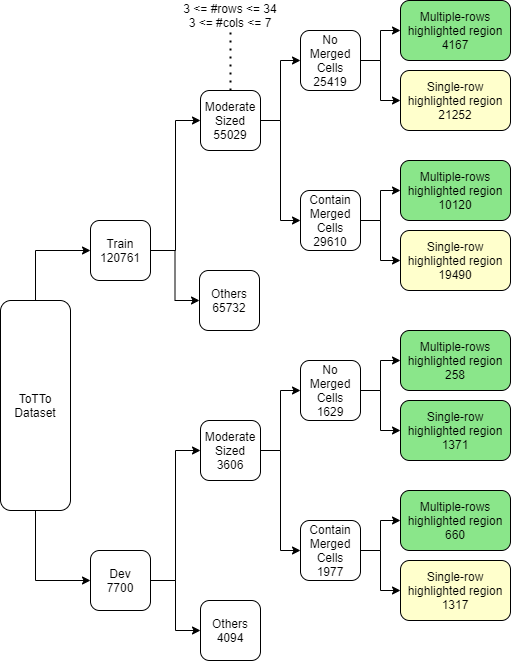}
  \caption{Flowchart of ToTTo filtering process}
  \label{fig:ToTTo_filtering_process}
\end{figure*}

\begin{figure*}[h]
  \centering
  \includegraphics[width=0.9\textwidth]{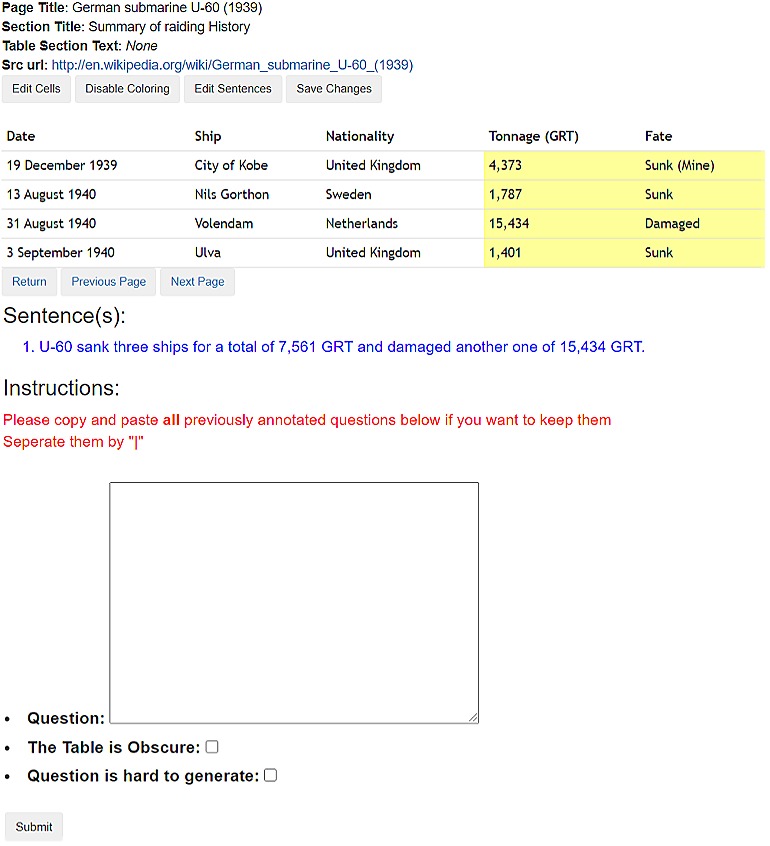}
  \caption{\ours annotation interface}
  \label{fig:website-screenshot}
\end{figure*}

\begin{figure*}[h]
  \centering
  \includegraphics[width=1\textwidth]{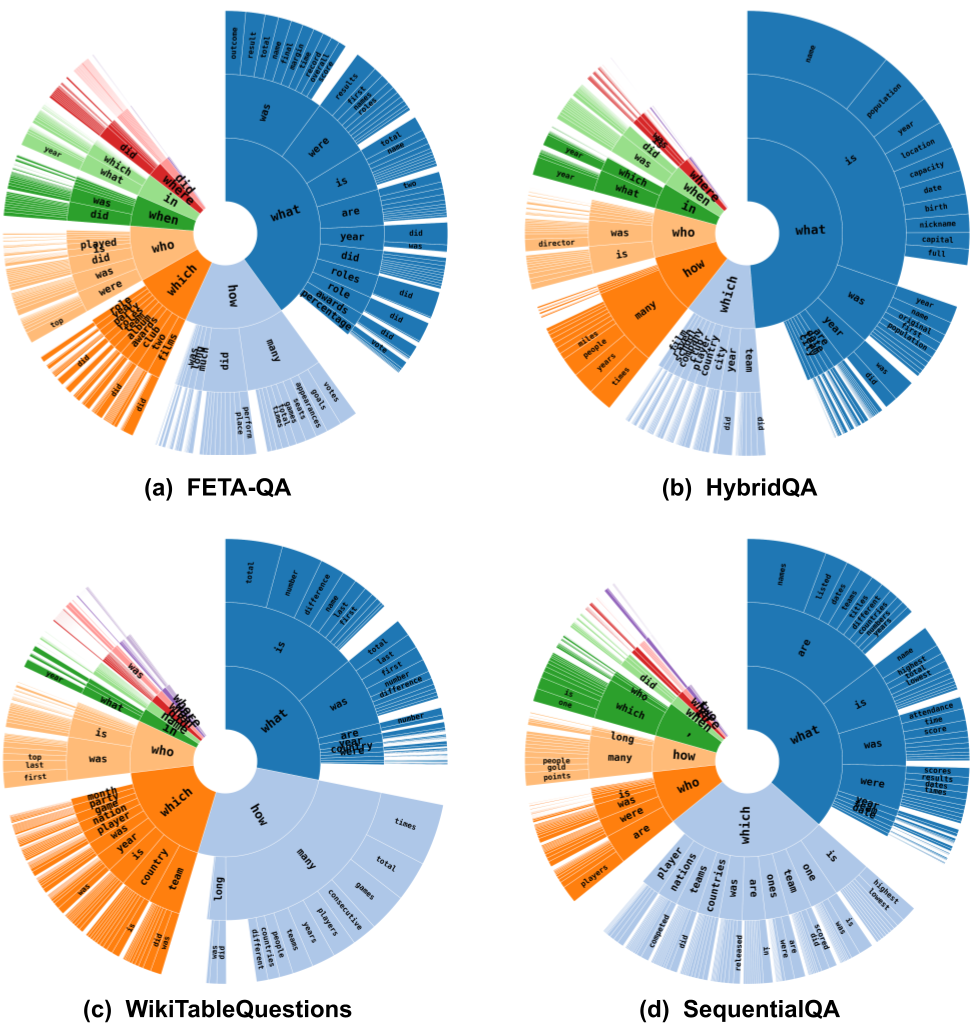}
  \caption{Question type distribution comparison between different Table QA datasets}
  \label{fig:all-question-types}
\end{figure*}

The Appendix contains the following contents:
\begin{itemize}
    \item Flowchart of ToTTo instances sampling process. (Figure \ref{fig:ToTTo_filtering_process})
    \item Screenshot of \ours annotation interface. (Figure \ref{fig:website-screenshot})
    \item Question type distribution comparison between \ours and other Table QA datasets. (Figure \ref{fig:all-question-types})
\end{itemize}

\end{document}